%% file: main.tex
\documentclass[10pt,twocolumn,letterpaper]{article}

\usepackage[pagenumbers]{cvpr} 

\usepackage{graphicx}
\usepackage{amsmath}
\usepackage{amssymb}
\usepackage{booktabs}
\usepackage{multirow}
\usepackage{pifont}
\usepackage{xcolor}
\def\etc{etc.\@\xspace}

\usepackage{colortbl}
\newcommand{\cc}{\cellcolor[gray]{0.9}}

\usepackage{listings}
\definecolor{codegreen}{rgb}{0,0.6,0}
\definecolor{codegray}{rgb}{0.5,0.5,0.5}
\definecolor{codepurple}{rgb}{0.58,0,0.82}

\lstdefinestyle{mystyle}{
    commentstyle=\color{codegreen},
    keywordstyle=\color{magenta},
    numberstyle=\tiny\color{codegray},
    stringstyle=\color{codepurple},
    basicstyle=\ttfamily\small,
    breakatwhitespace=false,         
    breaklines=true,                 
    captionpos=b,                    
    keepspaces=true,                 
    numbers=left,                    
    numbersep=5pt,                  
    showspaces=false,                
    showstringspaces=false,
    showtabs=false,                  
    tabsize=2
}
\usepackage[pagebackref,breaklinks,colorlinks]{hyperref}
\usepackage[accsupp]{axessibility}  

\usepackage[capitalize]{cleveref}
\crefname{section}{Sec.}{Secs.}
\Crefname{section}{Section}{Sections}
\Crefname{table}{Table}{Tables}
\crefname{table}{Tab.}{Tabs.}

\def\cvprPaperID{5440} 
\def\confName{CVPR}
\def\confYear{2023}

\begin{document}

\title{Run, Don't Walk: Chasing Higher FLOPS for Faster Neural Networks}
\author{Jierun Chen\textsuperscript{1}, Shiu-hong Kao\textsuperscript{1}, Hao He\textsuperscript{1} \\
Weipeng Zhuo\textsuperscript{1}, Song Wen\textsuperscript{2}, Chul-Ho Lee\textsuperscript{3}, S.-H. Gary Chan\textsuperscript{1}\\
\textsuperscript{1}HKUST,
\textsuperscript{2}Rutgers University,
\textsuperscript{3}Texas State University\\
}
\maketitle

\input{0_abstract}

\input{1_introduction}
\input{2_related_work}
\input{3_approach}
\input{4_experiments}
\input{5_conclusion}

\medskip\noindent\textbf{Acknowledgement} \enspace
This work was supported, in part, by Hong Kong General Research Fund under grant number 16200120. The work of C.-H. Lee was supported, in part, by the NSF under Grant IIS-2209921.
\input{6_appendix}

{\small
\bibliographystyle{ieee_fullname}
\bibliography{egbib}
}

\end{document}

%% file: 0_abstract.tex
\begin{abstract}
To design fast neural networks, many works have been focusing on reducing the number of floating-point operations (FLOPs). We observe that such reduction in FLOPs, however, does not necessarily lead to a similar level of reduction in latency. This mainly stems from inefficiently low floating-point operations per second (FLOPS). To achieve faster networks, we revisit popular operators and demonstrate that such low FLOPS is mainly due to frequent memory access of the operators, especially the depthwise convolution. We hence propose a novel partial convolution (PConv) that extracts spatial features more efficiently, by cutting down redundant computation and memory access simultaneously. Building upon our PConv, we further propose FasterNet, a new family of neural networks, which attains substantially higher running speed than others on a wide range of devices, without compromising on accuracy for various vision tasks. For example, on ImageNet-1k, our tiny FasterNet-T0 is $2.8\times$, $3.3\times$, and $2.4\times$ faster than MobileViT-XXS on GPU, CPU, and ARM processors, respectively, while being 2.9\% more accurate. Our large FasterNet-L achieves impressive 83.5\% top-1 accuracy, on par with the emerging Swin-B, while having 36\% higher inference throughput on GPU, as well as saving 37\% compute time on CPU. Code is available at \url{https://github.com/JierunChen/FasterNet}.
\end{abstract}

%% file: 1_introduction.tex
\section{Introduction}
\label{sec:intro}

Neural networks have undergone rapid development in various computer vision tasks such as image classification, detection and segmentation. While their impressive performance has powered many applications, a roaring trend is to pursue fast neural networks with low latency and high throughput for great user experiences, instant responses, safety reasons, etc.

How to be fast? Instead of asking for more costly computing devices, researchers and practitioners prefer to design cost-effective fast neural networks with reduced computational complexity, mainly measured in the number of {\bf fl}oating-point {\bf op}eration{\bf s} (FLOPs)\footnote{We follow a widely adopted definition of FLOPs, as the number of multiply-adds~\cite{zhang2018shufflenet,liu2022convnet}.}. MobileNets~\cite{howard2017mobilenets,sandler2018mobilenetv2,howard2019searching},
ShuffleNets~\cite{zhang2018shufflenet,ma2018shufflenet} and GhostNet~\cite{han2020ghostnet}, among others, leverage the depthwise convolution (DWConv)~\cite{sifre2014rigid} and/or group convolution (GConv)~\cite{krizhevsky2012imagenet} to extract spatial features. However, in the effort to reduce FLOPs, the operators often suffer from the side effect of increased memory access. MicroNet~\cite{li2021micronet} further decomposes and sparsifies the network to push its FLOPs to an extremely low level. Despite its improvement in FLOPs, this approach experiences inefficient fragmented computation. Besides, the above networks are often accompanied by additional data manipulations, such as concatenation, shuffling, and pooling, whose running time tends to be significant for tiny models.

\begin{figure}
    \centering
    \includegraphics[width=1\linewidth]{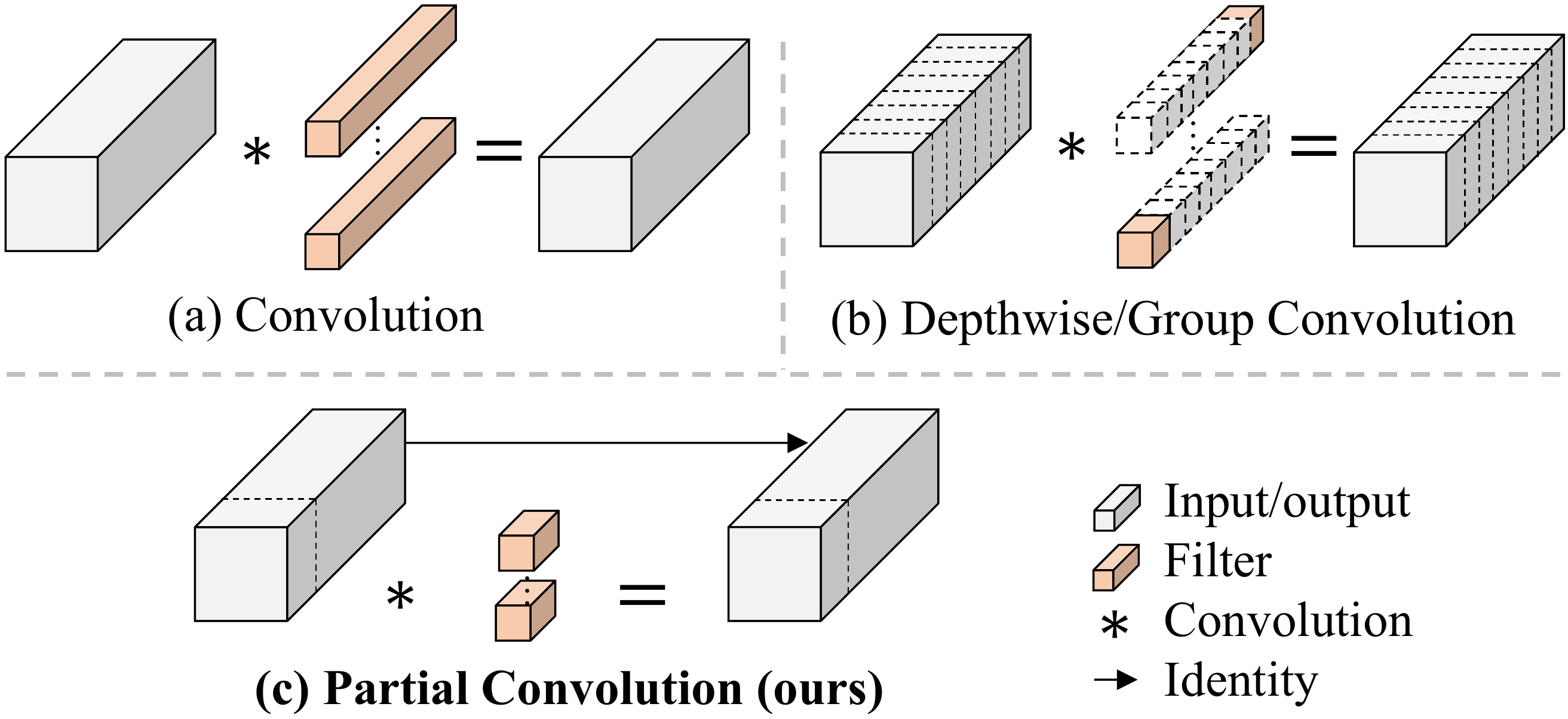}
    \vspace{-0.2in}
    \caption{Our partial convolution (PConv) is fast and efficient by applying filters on only a few input channels while leaving the remaining ones untouched. PConv obtains lower FLOPs than the regular convolution and higher FLOPS than the depthwise/group convolution.}
    \label{fig: PConv}
    \vspace{-0.05in}
\end{figure}

\begin{figure*}
    \centering
    \includegraphics[width=1\linewidth]{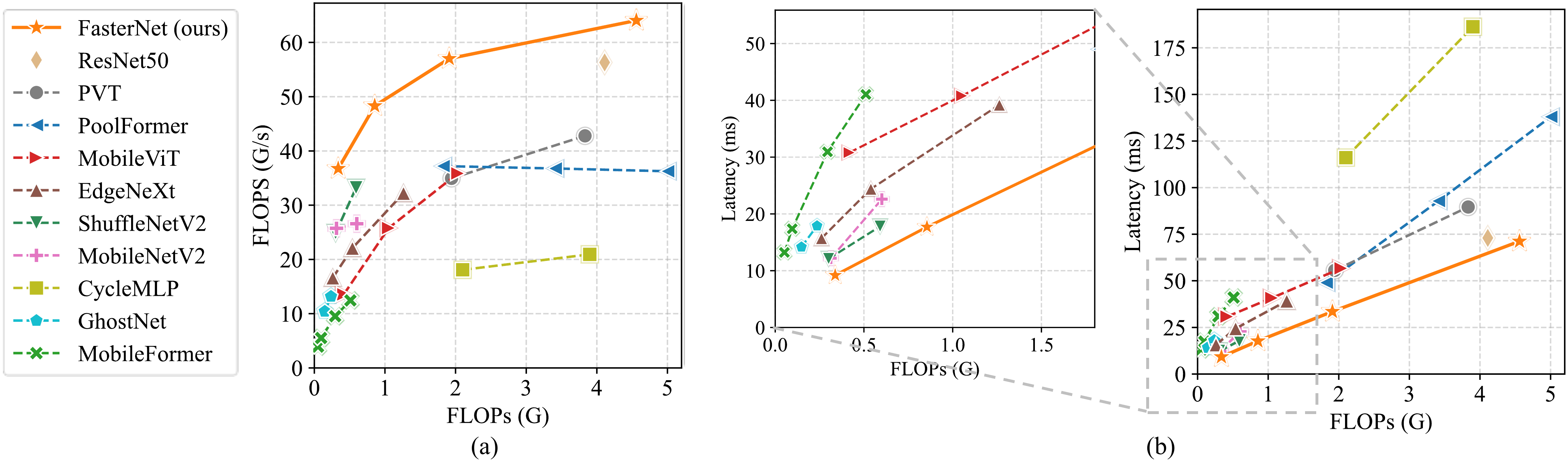}
    \vspace{-0.3in}
    \caption{(a) FLOPS under varied FLOPs on CPU. Many existing neural networks suffer from low computational speed issues. Their effective FLOPS are lower than the popular ResNet50. By contrast, our FasterNet attains higher FLOPS. (b) Latency under varied FLOPs on CPU. Our FasterNet obtains lower latency than others with the same amount of FLOPs.}
    \label{fig:FLOPS(latency)_vs_FLOPs}
    \vspace{-0.05in}
\end{figure*}

Apart from the above pure convolutional neural networks (CNNs),  there is an emerging interest in making vision transformers (ViTs)~\cite{dosovitskiy2020image} and multilayer perceptrons (MLPs) architectures~\cite{tolstikhin2021mlp} smaller and faster. For example, MobileViTs~\cite{mehta2021mobilevit,mehta2022separable,wadekar2022mobilevitv3} and MobileFormer~\cite{chen2022mobile} reduce the computational complexity by combining DWConv with a modified attention mechanism. However, they still suffer from the aforementioned issue with DWConv and also need dedicated hardware support for the modified attention mechanism. The use of advanced yet time-consuming normalization and activation layers may also limit their speed on devices.

All these issues together lead to the following question: Are
these ``fast'' neural networks really fast? To answer this, we examine the relationship between latency and FLOPs, which is captured by 
\begin{equation}
  Latency = \frac{FLOPs}{FLOPS},
  \label{eq:latency_FLOPs}
\end{equation}
where FLOPS is short for {\bf fl}oating-point {\bf op}erations per {\bf s}econd, as a measure of the effective computational speed. While there are many attempts to reduce FLOPs, they seldom consider optimizing FLOPS at the same time to achieve truly low latency. To better understand the situation, we compare the FLOPS of typical neural networks on an Intel CPU. The results in~\cref{fig:FLOPS(latency)_vs_FLOPs} show that many existing neural networks suffer from low FLOPS, and their FLOPS is generally lower than the popular ResNet50. With such low FLOPS, these ``fast'' neural networks are actually not fast enough.
Their reduction in FLOPs cannot be translated into the exact amount of reduction in latency. In some cases, there is no improvement, and it even leads to worse latency. For example, CycleMLP-B1~\cite{chen2021cyclemlp} has half of FLOPs of ResNet50~\cite{he2016deep} but runs more slowly (\ie, CycleMLP-B1 \vs ResNet50: 116.1ms \vs 73.0ms). Note that this discrepancy between FLOPs and latency has also been noticed in previous works~\cite{ma2018shufflenet,mehta2021mobilevit} but remains unresolved partially because they employ the DWConv/GConv and various data manipulations with low FLOPS. It is deemed there are no better alternatives available.

This paper aims to eliminate the discrepancy by developing a simple yet fast and effective operator that maintains high FLOPS with reduced FLOPs. Specifically, we reexamine existing operators, particularly  DWConv, in terms of the computational speed -- FLOPS. We uncover that the main reason causing the low FLOPS issue is \emph{frequent memory access}. We then propose a novel partial convolution (PConv) as a competitive alternative that reduces the computational redundancy as well as the number of memory access. \cref{fig: PConv} illustrates the design of our PConv. It takes advantage of redundancy within the feature maps and systematically applies a regular convolution (Conv) on only a part of the input channels while leaving the remaining ones untouched. By nature, PConv has lower FLOPs than the regular Conv while having higher FLOPS than the DWConv/GConv. In other words, PConv better exploits the on-device computational capacity. PConv is also effective in extracting spatial features as empirically validated later in the paper.

We further introduce FasterNet, which is primarily built upon our PConv, as a new family of networks that run highly fast on various devices. In particular, our FasterNet achieves state-of-the-art performance for classification, detection, and segmentation tasks while having much lower latency and higher throughput. For example, our tiny FasterNet-T0 is $2.8\times$, $3.3\times$, and $2.4\times$ faster than MobileViT-XXS~\cite{mehta2021mobilevit} on GPU, CPU, and ARM processors, respectively, while being 2.9\% more accurate on ImageNet-1k. Our large FasterNet-L achieves 83.5\% top-1 accuracy, on par with the emerging Swin-B~\cite{liu2021swin}, while offering 36\% higher throughput on GPU and saving 37\% compute time on CPU. To summarize, our contributions are as follows:
\begin{itemize}
\itemsep0em 
\item We point out the importance of achieving higher FLOPS beyond simply reducing FLOPs for faster neural networks.
\item We introduce a simple yet fast and effective operator called PConv, which has a high potential to replace the existing go-to choice, DWConv.
\item We introduce FasterNet which runs favorably and universally fast on a variety of devices such as GPU, CPU, and ARM processors.
\item We conduct extensive experiments on various tasks and validate the high speed and effectiveness of our PConv and FasterNet.
\end{itemize}

%% file: 2_related_work.tex
\section{Related Work}
\label{sec:related_work}
We briefly review prior works on fast and efficient neural networks and differentiate this work from them.

\medskip\noindent\textbf{CNN.} \enspace
CNNs are the mainstream architecture in the computer vision field, especially when it comes to deployment in practice, where being fast is as important as being accurate. Though there have been numerous studies~\cite{sifre2014rigid,singh2019hetconv,chen2019drop,chollet2017xception,zhang2017interleaved,li2021micronet,he2022tackling,zhuo2022semi} to achieve higher efficiency, the rationale behind them is more or less to perform a low-rank approximation. Specifically, the group convolution~\cite{krizhevsky2012imagenet} and the depthwise separable convolution~\cite{sifre2014rigid} (consisting of depthwise and pointwise convolutions) are probably the most popular ones. They have been widely adopted in mobile/edge-oriented networks, such as MobileNets~\cite{howard2017mobilenets,sandler2018mobilenetv2,howard2019searching},
ShuffleNets~\cite{zhang2018shufflenet,ma2018shufflenet}, GhostNet~\cite{han2020ghostnet},
EfficientNets~\cite{tan2019efficientnet,tan2021efficientnetv2}, TinyNet~\cite{han2020model}, Xception~\cite{chollet2017xception}, CondenseNet~\cite{huang2018condensenet,yang2021condensenet}, TVConv~\cite{chen2022tvconv}, MnasNet\cite{tan2019mnasnet}, and FBNet~\cite{wu2019fbnet}. While they exploit the redundancy in filters to reduce the number of parameters and FLOPs, they suffer from increased memory access when increasing the network width to compensate for the accuracy drop. By contrast, we consider the redundancy in feature maps and propose a partial convolution to reduce FLOPs and memory access \emph{simultaneously}. 

\medskip\noindent\textbf{ViT, MLP, and variants.} \enspace
 There is a growing interest in studying ViT ever since Dosovitskiy \etal~\cite{dosovitskiy2020image} expanded the application scope of transformers~\cite{vaswani2017attention} from machine translation~\cite{vaswani2017attention} or forecasting~\cite{wen2022social} to the computer vision field. Many follow-up works have attempted to improve ViT in terms of training setting~\cite{touvron2021training,touvron2022deit,steiner2021train} and model design~\cite{liu2021swin,liu2022swin,wang2021pyramid,graham2021levit,zhong2022tree}. One notable trend is to pursue a better accuracy-latency trade-off by reducing the complexity of the attention operator~\cite{ali2021xcit,vaswani2021scaling,huang2022lightvit,lu2021soft,tang2022quadtree}, incorporating convolution into ViTs~\cite{dai2021coatnet,chen2022mobile,srinivas2021bottleneck}, or doing both~\cite{cai2022efficientvit,li2022efficientformer,pan2022edgevits,mehta2022separable}. Besides, other studies~\cite{tolstikhin2021mlp,lian2021mlp,chen2021cyclemlp} propose to replace the attention with simple MLP-based operators. However, they often evolve to be CNN-like~\cite{liu2022we}. In this paper, we focus on analyzing the convolution operations, particularly DWConv, due to the following reasons: First, the advantage of attention over convolution is unclear or debatable~\cite{wang2022shift,liu2022convnet}. Second, the attention-based mechanism generally runs slower than its convolutional counterparts and thus  becomes less favorable for the current industry~\cite{mehta2021mobilevit,hu2019local}. Finally, DWConv is still a popular choice in many hybrid models, so it is worth a careful examination.
 

%% file: 3_approach.tex
\section{Design of PConv and FasterNet}
\label{sec:approach}
In this section, we first revisit DWConv and analyze the issue with its frequent memory access. We then introduce PConv as a competitive alternative operator to resolve the issue. After that, we introduce FasterNet and explain its details, including design considerations. 

\subsection{Preliminary}
DWConv is a popular variant of Conv and has been widely adopted as a key building block for many neural networks. For an input 
$ \mathbf{I} \in \mathbb{R}^{ c \times h \times w}$, DWConv applies 
$c$
filters
$ \mathbf{W} \in \mathbb{R}^{ k \times k} $
to compute the output 
$ \mathbf{O} \in \mathbb{R}^{ c \times h \times w}$.
As shown in~\cref{fig: PConv}(b), each filter slides spatially on one input channel and contributes to one output channel.
This depthwise computation makes DWConv have as low FLOPs as $ h \times w \times k^2 \times c$ compared to a regular Conv with
$ h \times w \times k^2 \times c^2$. While effective in reducing FLOPs, a DWConv, which is typically followed by a pointwise convolution, or PWConv, cannot be simply used to replace a regular Conv as it would incur a severe accuracy drop.
Thus, in practice the channel number
$c$ (or the network width) of DWConv is increased to 
$c'\left(c' > c\right)$
to compensate the accuracy drop, \eg, the width is expanded by six
times for the DWConv in the inverted residual blocks~\cite{sandler2018mobilenetv2}. This, however, results in much higher memory access that can cause non-negligible delay and slow down the overall computation, especially for I/O-bound devices. In particular, the number of memory access now escalates to 
\begin{equation}
  h \times w \times 2c' + k^2 \times c'\approx h \times w \times 2c',
  \label{eq:memory_access_DWConv}
\end{equation}
which is higher than that of a regular Conv, \ie,
\begin{equation}
  h \times w \times 2c + k^2 \times c^2 \approx h \times w \times 2c.
  \label{eq:memory_access_Conv}
\end{equation}
Note that the $h \times w \times 2c'$ memory access is spent on the I/O operation, which is deemed to be already the minimum cost and hard to optimize further.

\begin{figure}
    \centering
    \includegraphics[width=1\linewidth]{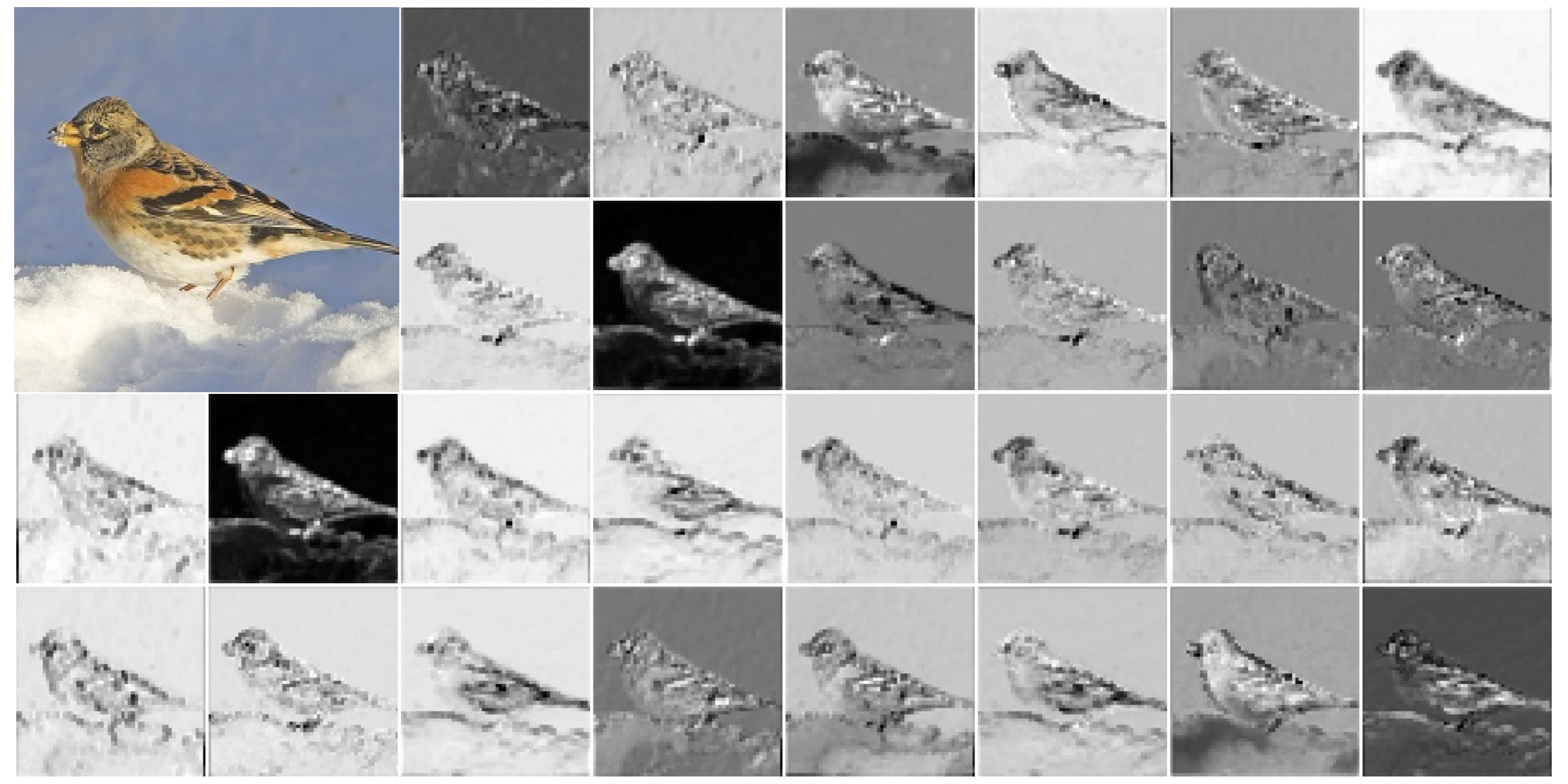}
    \vspace{-0.25in}
    \caption{Visualization of feature maps in an intermediate layer of a pre-trained ResNet50, with the top-left image as the input. Qualitatively, we can see the high redundancies across different channels. }
    \label{fig: feature_redundancy}
    \vspace{-0.05in}
\end{figure}

\subsection{Partial convolution as a basic operator}
We below demonstrate that the cost can be further optimized by leveraging the feature maps' redundancy. As visualized in~\cref{fig: feature_redundancy}, the feature maps share high similarities among different channels. This redundancy has also been covered in many other works~\cite{han2020ghostnet,zhang2020split}, but few of them make full use of it in a simple yet effective way.

\begin{figure*}
    \vspace{-0.05in}
    \centering
    \includegraphics[width=0.83\linewidth]{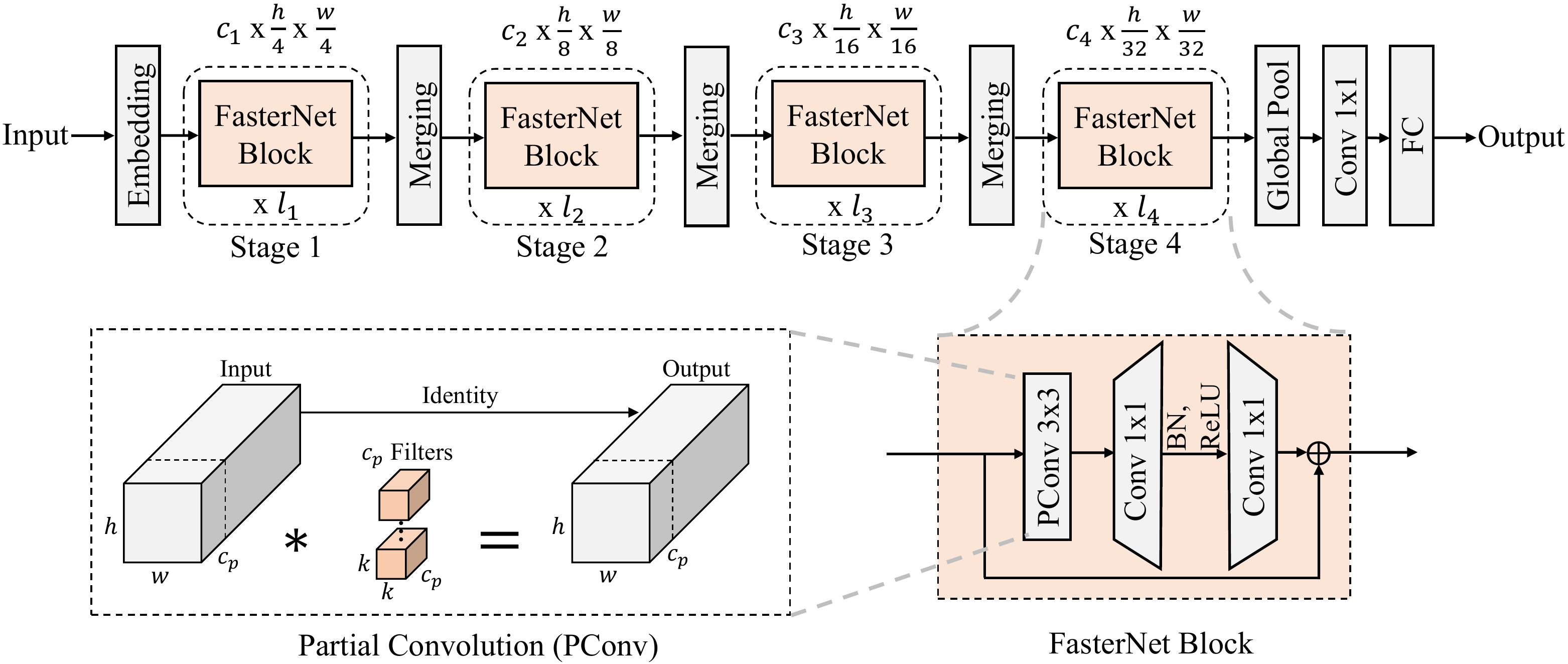}
    \vspace{-0.1in}
    \caption{Overall architecture of our FasterNet. It has four hierarchical stages, each with a stack of FasterNet blocks and preceded by an embedding or merging layer. The last three layers are used for feature classification. Within each FasterNet block, a PConv layer is followed by two PWConv layers. We put normalization and activation layers only after the middle layer to preserve the feature diversity and achieve lower latency.
    }
    \label{fig: FasterNet}
    \vspace{-0.15in}
\end{figure*}

\begin{figure}
    \centering
    \includegraphics[width=.92\linewidth]{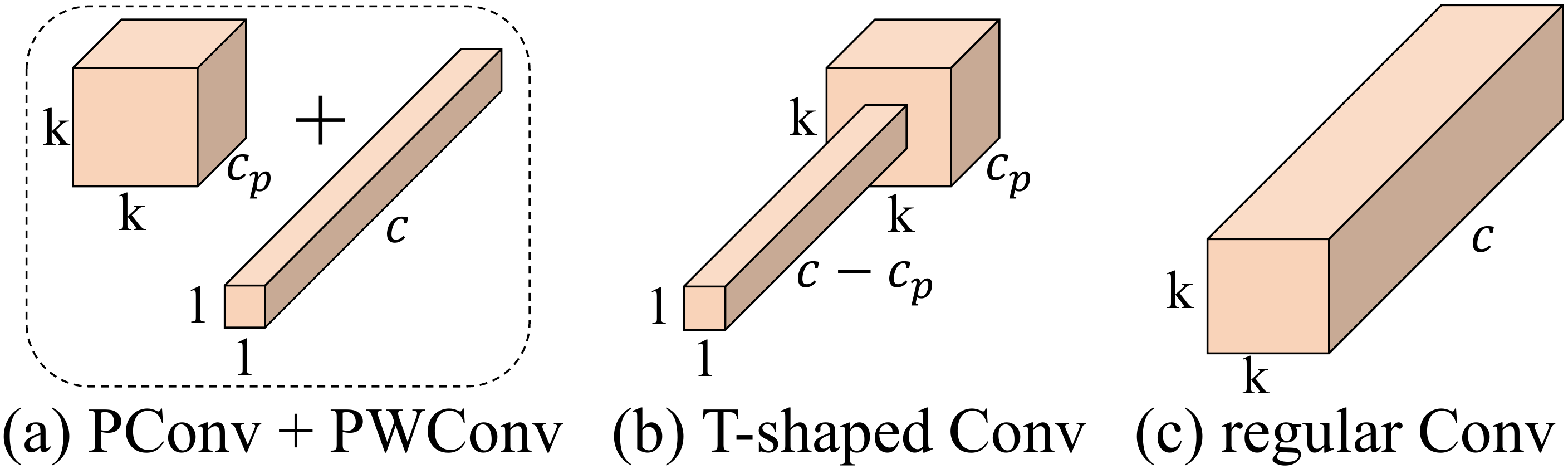}
    \vspace{-0.1in}
    \caption{Comparison of convolutional variants. A PConv followed by a PWConv (a) resembles a 
    T-shaped
     Conv (b), which spends more computation on the center position compared to a regular Conv (c).}
    \label{fig: pconv_pwconv}
    \vspace{-0.1in}
\end{figure}

Specifically, we propose a simple PConv to reduce computational redundancy and memory access simultaneously. The bottom-left corner in~\cref{fig: FasterNet} illustrates how our PConv works. It simply applies a regular Conv on only a part of the input channels for spatial feature extraction and leaves the remaining channels untouched. For contiguous or regular memory access, we consider the first or last consecutive $c_p$
channels as the representatives of the whole feature maps for computation. Without loss of generality, we consider the input and output feature maps to have the same number of channels. Therefore, the FLOPs of a PConv are only 
\begin{equation}
   h \times w \times k^2 \times c_p^2.
  \label{eq:FLOPs_PConv}
\end{equation}
With a typical partial ratio $r \!=\! \frac{c_p}{c} \!=\! \frac{1}{4}$,
the FLOPs of a PConv is only $\frac{1}{16}$
of a regular Conv. Besides, PConv has a smaller amount of memory access, \ie,
\begin{equation}
  h \times w \times 2c_p + k^2 \times c_p^2 \approx h \times w \times 2c_p,
  \label{eq:memory_access_PConv}
\end{equation}
which is only 
$\frac{1}{4}$
of a regular Conv for 
$r=\frac{1}{4}$. 

Since there are only $c_p$
channels utilized for spatial feature extraction, one may ask if we can simply remove the remaining $(c-c_p)$
channels? If so, PConv would degrade to a regular Conv with fewer channels, which deviates from our objective to reduce redundancy. Note that we keep the remaining channels untouched instead of removing them from the feature maps. It is because they are useful for a subsequent PWConv layer, which allows the feature information to flow through all channels.

\subsection{PConv followed by PWConv }
\begin{figure}
    \vspace{-0.15in}
    \centering
    \includegraphics[width=.99\linewidth]{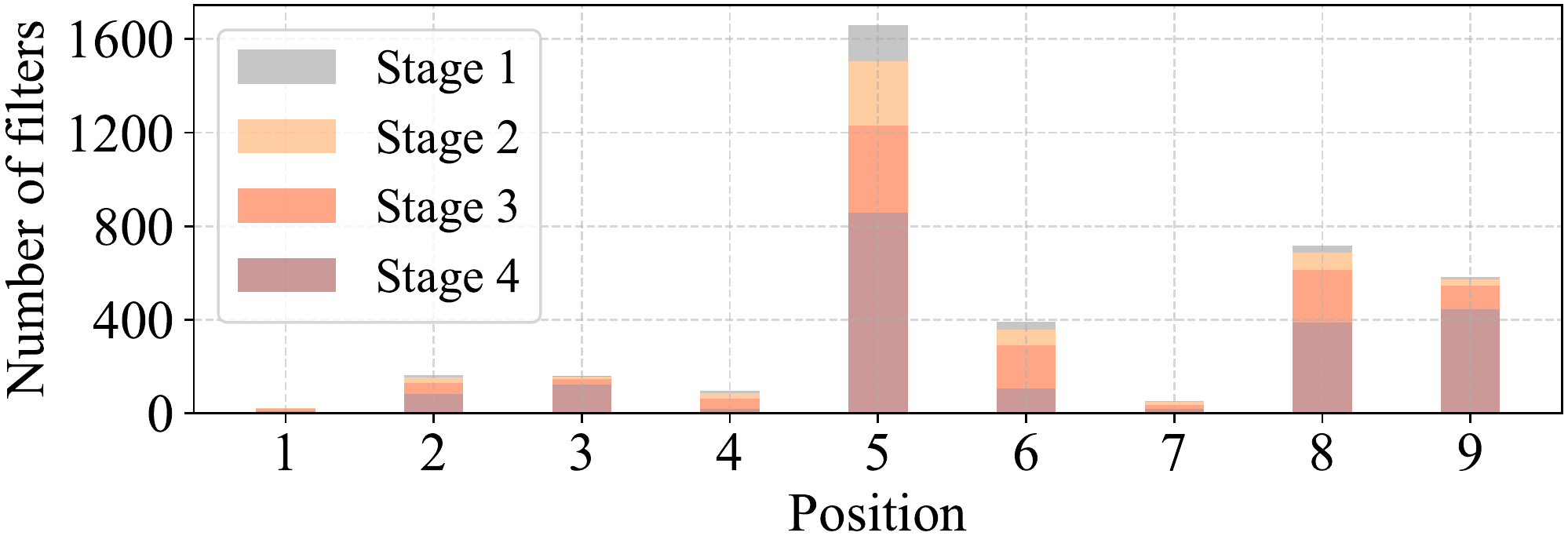}
    \vspace{-0.1in}
    \caption{Histogram of salient position distribution for the regular Conv 
    $3 \times 3$
    filters in a pre-trained ResNet18. The histogram contains four kinds of bars, corresponding to different stages in the network. In all stages, the center position (position 5) appears as a salient position most frequently.}
    \label{fig: positionwise_norm}
    \vspace{-0.1in}
\end{figure}

To fully and efficiently leverage the information from all channels, we further append a pointwise convolution (PWConv) to our PConv. Their effective receptive field together on the input feature maps looks like a 
T-shaped
Conv, which focuses more on the center position compared to a regular Conv uniformly processing a patch, as shown in~\cref{fig: pconv_pwconv}.
To justify this T-shaped
receptive field, we first evaluate the importance of each position by calculating the position-wise Frobenius norm. We assume that a position tends to be more important if it has a larger Frobenius norm than other positions. 
For a regular Conv filter
$\mathbf{F} \in \mathbb{R}^{  k^2 \times  c}$, 
the Frobenius norm at position 
$i$
is calculated by
$\left\|\mathbf{F}_{i}\right\| \!=\! \sqrt{
  \sum_{j=1}^c |f_{ij}|^2
  },
$
for $i = 1, 2, 3 ..., k^2$.
We consider a salient position to be the one with the maximum Frobenius norm.
We then collectively examine each filter in a pre-trained ResNet18, find out their salient positions, and plot a histogram of the salient positions. 
Results in~\cref{fig: positionwise_norm} show that the center position turns out to be the salient position most frequently among the filters. 
In other words, the center position weighs more than its surrounding neighbors. This is consistent with the T-shaped computation which concentrates on the center position.
 
While the
 T-shaped Conv can be directly used for efficient computation, we show that it is better to decompose the T-shaped Conv into a PConv and a PWConv because the decomposition exploits the inter-filter redundancy and  further saves FLOPs. For the same input 
 $\mathbf{I} \in \mathbb{R}^{ c \times h \times w}$ 
 and output 
 $\mathbf{O} \in \mathbb{R}^{ c \times h \times w}$,
 a T-shaped Conv's FLOPs can be calculated as 
 \begin{equation}
h \times w \times \left(k^2 \times c_p \times c + c \times \left(c - c_p\right) \right),
  \label{eq:FLOPs_tuConv}
\end{equation}
which is higher than the FLOPs of a PConv and a PWConv, \ie,
\begin{equation}
   h \times w \times ( k^2 \times c_p^2 + c^2),
  \label{eq:FLOPs_PConv_PWConv}
\end{equation}
where
$
(k^{2} - 1)c > k^{2} c_p,
$
\eg when 
$
c_p = \frac{c}{4}
$
and 
$
k = 3.
$
Besides, we can readily leverage the regular Conv for the two-step implementation.

\subsection{FasterNet as a general backbone}
Given our novel PConv and off-the-shelf PWConv as the primary building operators, we further propose FasterNet, a new family of neural networks that runs favorably fast and is highly effective for many vision tasks. We aim to keep the architecture as simple as possible, without bells and whistles, to make it hardware-friendly in general. 

We present the overall architecture in~\cref{fig: FasterNet}. It has four hierarchical stages, each of which is preceded by an embedding layer (a regular Conv $4 \times 4$ with stride 4) or a merging layer (a regular Conv $2 \times 2$ with stride 2) for spatial downsampling and channel number expanding. 
Each stage has a stack of FasterNet blocks. We observe that the blocks in the last two stages consume less memory access and tend to have higher FLOPS, as empirically validated in~\cref{FLOPS_compare}. Thus, we put more FasterNet blocks and correspondingly assign more computations to the last two stages. Each FasterNet block has a PConv layer followed by two PWConv (or Conv $1 \times 1$) layers. Together, they appear as inverted residual blocks where the middle layer has an expanded number of channels, and a shortcut connection is placed to reuse the input features. 

In addition to the above operators, the normalization and activation layers are also indispensable for high-performing neural networks.
Many prior works~\cite{he2016deep,sandler2018mobilenetv2,han2020ghostnet}, however, overuse such layers throughout the network, which may limit the feature diversity and thus hurt the performance. It can also slow down the overall computation. By contrast, we put them only after each middle PWConv to preserve the feature diversity and achieve lower latency.
Besides, we use the batch normalization (BN)~\cite{ioffe2015batch} instead of other alternative ones~\cite{ba2016layer,ulyanov2016instance,wu2018group}. The benefit of BN is that it can be merged into its adjacent Conv layers for faster inference while being as effective as the others. As for the activation layers, we empirically choose GELU~\cite{hendrycks2016gaussian} for smaller FasterNet variants and ReLU~\cite{nair2010rectified} for bigger FasterNet variants, considering both running time and effectiveness. The last three layers, \ie a global average pooling, a Conv $1 \times 1$, and a fully-connected layer, are used together for feature transformation and classification.

To serve a wide range of applications under different computational budgets, we provide tiny, small, medium, and large variants of FasterNet, referred to as FasterNet-T0/1/2, FasterNet-S, FasterNet-M, and FasterNet-L, respectively. They share a similar architecture but vary in depth and width. Detailed architecture specifications are provided in the appendix.

%% file: 4_experiments.tex
\section{Experimental Results}
\label{sec:experiments}
We first examine the computational speed of our PConv and its effectiveness when combined with a PWConv. We then comprehensively evaluate the performance of our FasterNet for classification, detection, and segmentation tasks. Finally, we conduct a brief ablation study. 

To benchmark the latency and throughput, we choose the following three typical processors, which cover a wide range of computational capacity: GPU (2080Ti), CPU (Intel i9-9900X, using a single thread), and ARM (Cortex-A72, using a single thread). We report their latency for inputs with a batch size of 1 and throughput for inputs with a batch size of 32. During inference, the BN layers are merged to their adjacent layers wherever applicable.

\subsection{PConv is fast with high FLOPS}
\input{tables/compare_ope0}
We below show that our PConv is fast and better exploits the on-device computational capacity. Specifically, we stack 10 layers of pure PConv and take feature maps of typical dimensions as inputs. We then measure FLOPs and latency/throughput on GPU, CPU, and ARM processors, which also allow us to further compute FLOPS. We repeat the same procedure for other convolutional variants and make comparisons. 

Results in~\cref{FLOPS_compare} show that PConv is overall an appealing choice for high FLOPS with reduced FLOPs. It has only $\frac{1}{16}$ FLOPs of a regular Conv and achieves $10.5\times$, $6.2\times$, and $22.8\times$ higher FLOPS than the DWConv on GPU, CPU, and ARM, respectively. We are unsurprised to see that the regular Conv has the highest FLOPS as it has been constantly optimized for years. However, its total FLOPs and latency/throughput are unaffordable. GConv and DWConv, despite their significant reduction in FLOPs, suffer from a drastic decrease in FLOPS. In addition, they tend to increase the number of channels to compensate for the performance drop, which, however, increase their latency.

\subsection{PConv is effective together with PWConv}
We next show that a PConv followed by a PWConv is effective in approximating a regular Conv to transform the feature maps. To this end, we first build four datasets by feeding the ImageNet-1k val split images into a pre-trained ResNet50, and extract the feature maps before and after the first Conv $3\times3$ in each of the four stages. Each feature map dataset is further spilt into the train (70\%), val (10\%), and test (20\%) subsets. We then build a simple network consisting of a PConv followed by a PWConv and train it on the feature map datasets with a mean squared error loss. For comparison, we also build and train networks for DWConv + PWConv and GConv + PWConv under the same setting.

\cref{tab:PConvPWConv} shows that PConv + PWConv achieve the lowest test loss, meaning that they better approximate a regular Conv in feature transformation. The results also suggest that it is sufficient and efficient to capture spatial features from only a part of the feature maps. PConv shows a great potential to be the new go-to choice in designing fast and effective neural networks.

\input{tables/pconv_pwconv}

\subsection{FasterNet on ImageNet-1k classification}

\begin{figure}
    \centering
    \vspace{-0.05in}
    \includegraphics[width=1\linewidth]{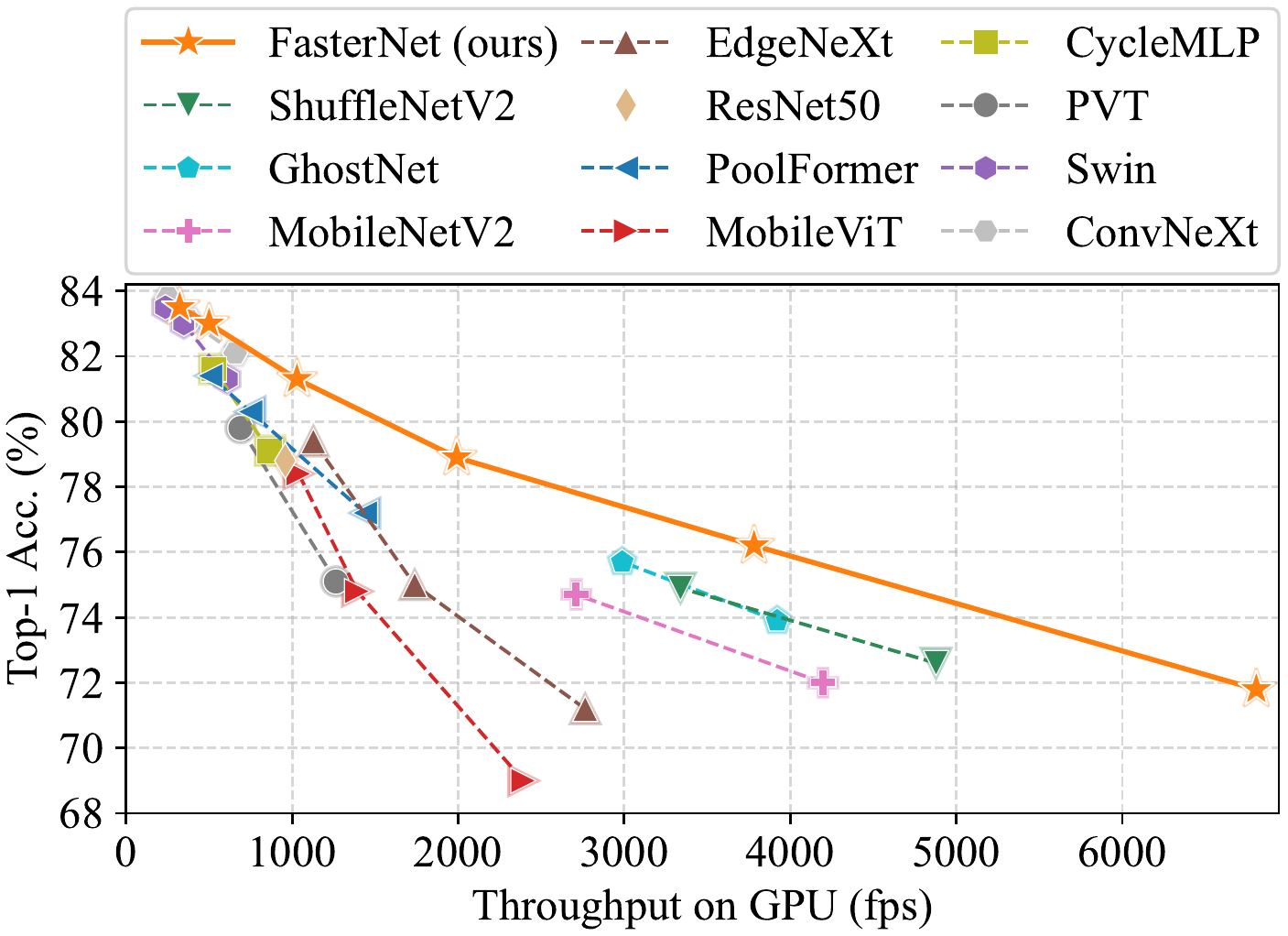}
    
    \includegraphics[width=1\linewidth]{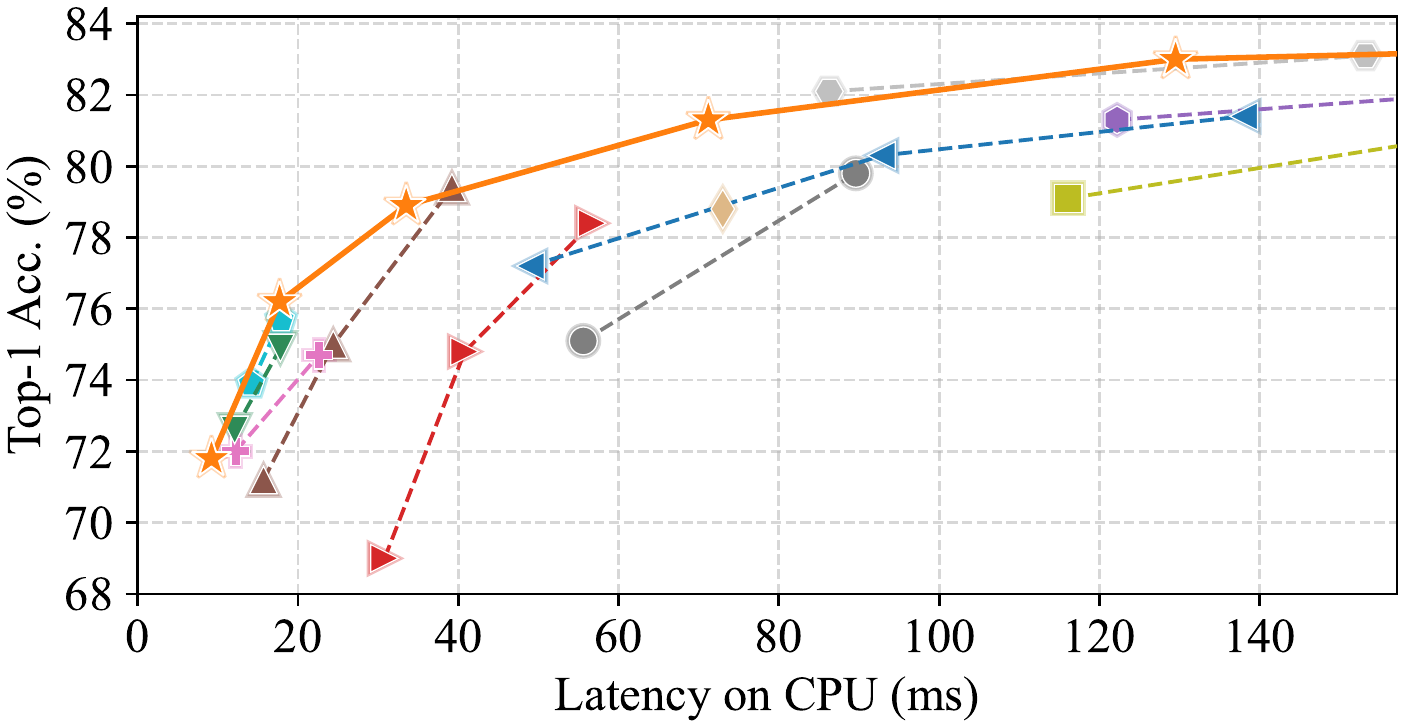}
    
    \includegraphics[width=1\linewidth]{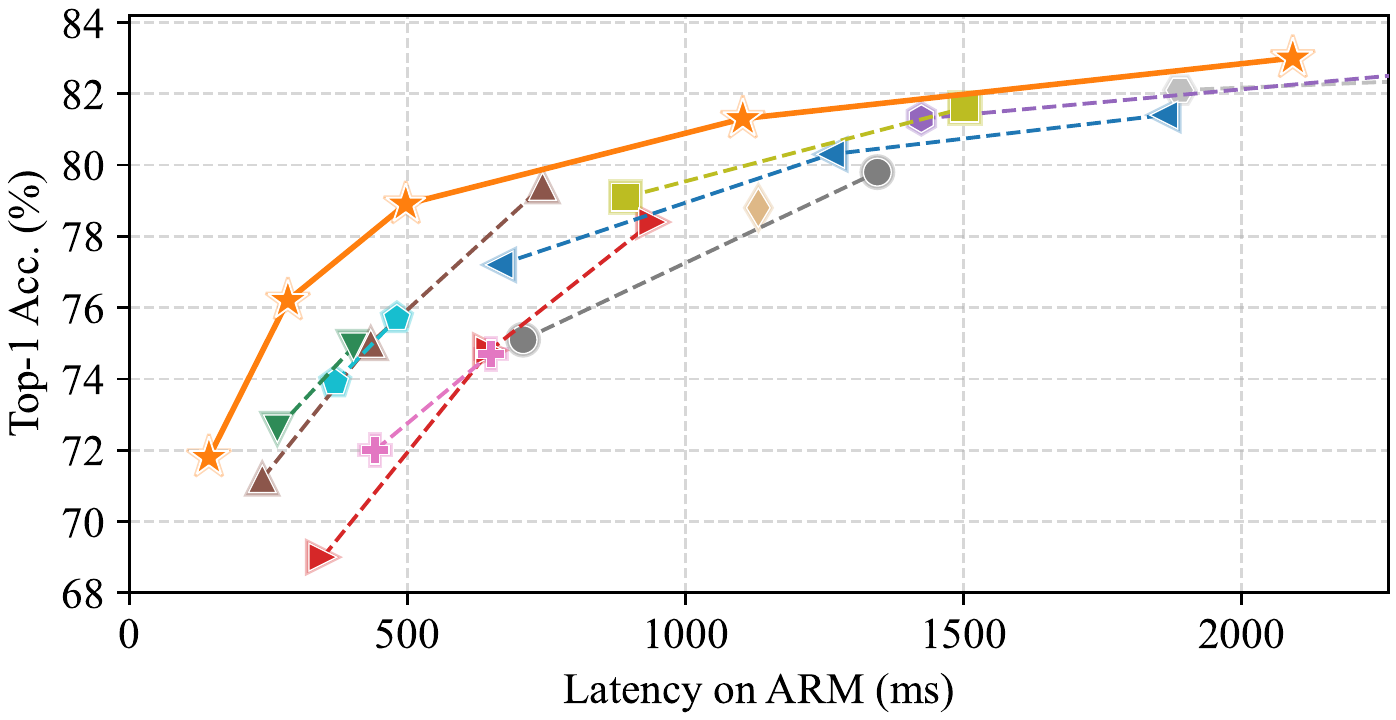}
    
    \vspace{-0.15in}
    \caption{FasterNet has the highest efficiency in balancing accuracy-throughput and accuracy-latency trade-offs for different devices. To save space and make the plots more proportionate, we showcase network variants within a certain range of latency. Full plots can be found in the appendix, which show consistent results.}
    \label{fig:imageent}
    \vspace{-0.15in}
\end{figure}

\input{tables/imagenet.tex}
\input{tables/coco.tex}
\input{tables/ablation.tex}

To verify the effectiveness and efficiency of our FasterNet, we first conduct experiments on the large-scale ImageNet-1k classification dataset~\cite{russakovsky2015imagenet}. It covers 1k categories of common objects and contains about 1.3M labeled images for training and 50k labeled images for validation. We train our models for 300 epochs using AdamW optimizer~\cite{loshchilov2017decoupled}. We set the batch size to 2048 for the FasterNet-M/L and 4096 for other variants. We use cosine learning rate scheduler~\cite{loshchilov2016sgdr} with a peak value of $0.001\cdot\text{batch size}/1024$ and a 20-epoch linear warmup. We apply commonly-used regularization and augmentation techniques, including Weight Decay~\cite{krogh1991simple}, Stochastic Depth~\cite{huang2016deep}, Label Smoothing~\cite{szegedy2016rethinking}, Mixup~\cite{zhang2017mixup}, Cutmix~\cite{yun2019cutmix} and Rand Augment~\cite{cubuk2020randaugment}, with varying magnitudes for different FasterNet variants. To reduce the training time, we use $192 \times 192$ resolution for the first 280 training epochs and $224 \times 224$ for the remaining 20 epochs. For fair comparison, we do not use knowledge distillation~\cite{hinton2015distilling} and neural architecture search~\cite{zoph2016neural}. We report our top-1 accuracy on the validation set with a center crop at $224 \times 224$ resolution and a 0.9 crop ratio. Detailed training and validation settings are provided in the appendix.

\cref{fig:imageent} and \cref{tab:imagenet} demonstrate the superiority of our FasterNet over state-of-the-art classification models. The trade-off curves in~\cref{fig:imageent} clearly show that FasterNet sets the new state-of-the-art in balancing accuracy and latency/throughput among all the networks examined. From another perspective, FasterNet runs faster than various CNN, ViT and MLP models on a wide range of devices, when having similar top-1 accuracy. As quantitatively shown in~\cref{tab:imagenet}, FasterNet-T0 is $2.8\times$, $3.3\times$, and $2.4\times$
faster than MobileViT-XXS~\cite{mehta2021mobilevit} on GPU, CPU, and ARM processors, respectively, while being 2.9\% more accurate.
Our large FasterNet-L achieves 83.5\% top-1 accuracy, comparable to the emerging Swin-B~\cite{liu2021swin} and ConvNeXt-B~\cite{liu2022convnet} while having 36\% and 28\% higher inference throughput on GPU, as well as saving 37\% and 15\% compute time on CPU. Given such promising results, we highlight that our FasterNet is much simpler than many other models in terms of architectural design, which showcases the feasibility of designing simple yet powerful neural networks.

\subsection{FasterNet on downstream tasks}
To further evaluate the generalization ability of FasterNet, we conduct experiments on the challenging COCO dataset~\cite{lin2014microsoft} for object detection and instance segmentation. As a common practice, we employ the ImageNet pre-trained FasterNet as a backbone and equip it with the popular Mask R-CNN detector~\cite{he2017mask}. To highlight the effectiveness of the backbone itself, we simply follow PoolFormer~\cite{yu2022metaformer} and adopt an AdamW optimizer, a $1 \times$ training schedule (12 epochs), a batch size of 16, and other training settings without further hyper-parameter tuning.

\cref{tab:coco} shows the results for comparison between FasterNet and representative models. FasterNet consistently outperforms ResNet and ResNext by having higher average precision (AP) with similar latency. Specifically, FasterNet-S yields $+1.9$ higher box AP and $+2.4$ higher mask AP compared to the standard baseline ResNet50. FasterNet is also competitive against the ViT variants. Under similar FLOPs, FasterNet-L reduces PVT-Large's latency by 38\%, \ie, from 152.2 ms to 93.8 ms on GPU, and achieves $+1.1$ higher box AP and $+0.4$ higher mask AP.

\subsection{Ablation study}
We conduct a brief ablation study on the value of partial ratio $r$ and the choices of activation and normalization layers. We compare different variants in terms of ImageNet top-1 accuracy and on-device latency/throughput. Results are summarized in~\cref{tab:ablation}. For the partial ratio $r$, we set it to $\frac{1}{4}$ for all FasterNet variants by default, which achieves higher accuracy, higher throughput, and lower latency at similar complexity. A too large partial ratio $r$ would make PConv degrade to a regular Conv, while a too small value would render PConv less effective in capturing the spatial features. 
For the normalization layers, we choose BatchNorm over LayerNorm because BatchNorm can be merged into its adjacent convolutional layers for faster inference while it is as effective as LayerNorm in our experiment.
For the activation function, interestingly, we empirically found that GELU fits FasterNet-T0/T1 models more efficiently than ReLU. It, however, becomes opposite for FasterNet-T2/S/M/L. Here we only show two examples in~\cref{tab:ablation} due to space constraint. We conjecture that GELU strengthens FasterNet-T0/T1 by having higher non-linearity, while the benefit fades away for larger FasterNet variants.

%% file: tables/compare_ope0.tex
\begin{table*}
\centering
\resizebox{.82\textwidth}{!}{%
\setlength{\tabcolsep}{2pt}
\begin{tabular}{c|cc|cc|cc|cc}
\toprule
\multirow{2}{*}{\begin{tabular}[c]{@{}c@{}}Operator\end{tabular}} & \multirow{2}{*}{\begin{tabular}[c]{@{}c@{}}Feature map\\ size\end{tabular}} & \multirow{2}{*}{\begin{tabular}[c]{@{}c@{}}FLOPs (M), \\ $\times$10 layers\end{tabular}} & \multicolumn{2}{c|}{GPU} & \multicolumn{2}{c|}{CPU} & \multicolumn{2}{c}{ARM} \\
 &  &  & \small Throughput (fps) & \small FLOPS (G/s) & \small Latency (ms) & \small FLOPS (G/s) & \small Latency (ms) & \small FLOPS (G/s) \\ \midrule
\multirow{5}{*}{Conv 3$\times$3} & 96$\times$56$\times$56 & 2601 & 3010 & 7824 & 35.67 & 72.90 & 779.57 & 3.33 \\
 & 192$\times$28$\times$28 & 2601 & 4893 & 12717 & 28.41 & 91.53 & 619.64 & 4.19 \\
 & 384$\times$14$\times$14 & 2601 & 4558 & 11854 & 31.85 & 81.66 & 595.09 & 4.37 \\
 & 768$\times$7$\times$7 & 2601 & 3159 & 8212 & 62.71 & 41.47 & 662.17 & 3.92 \\ \cline{2-9} 
 &\multicolumn{2}{c|}{\cc Average} &\cc - & \cc{10151} &\cc - & \cc{71.89} &\cc - & \cc{3.95} \\ \midrule 
\multirow{5}{*}{\begin{tabular}[c]{@{}c@{}}GConv 3$\times$3 \\ (16 groups)\end{tabular}} & 96$\times$56$\times$56 & 162 & 2888 & 469 & 21.90 & 7.42 & 166.30 & 0.97 \\
 & 192$\times$28$\times$28 & 162 & 10811 & 1754 & 7.58 & 21.44 & 96.22 & 1.68 \\
 & 384$\times$14$\times$14 & 162 & 15534 & 2514 & 4.40 & 36.88 & 63.57 & 2.55 \\
 & 768$\times$7$\times$7 & 162 & 16000 & 2598 & 4.28 & 37.97 & 65.20 & 2.49 \\ \cline{2-9} 
 &\multicolumn{2}{c|}{\cc Average} &\cc - & \cc{1833} &\cc - & \cc{25.93} &\cc - & \cc{1.92} \\ \midrule 
\multirow{5}{*}{\begin{tabular}[c]{@{}c@{}}DWConv 3$\times$3\end{tabular}} & 96$\times$56$\times$56 & 27.09 & 11940 & 323 & 3.59 & 7.52 & 108.70 & 0.24 \\
 & 192$\times$28$\times$28 & 13.54 & 23358 & 315 & 1.97 & 6.86 & 82.01 & 0.16 \\
 & 384$\times$14$\times$14 & 6.77 & 46377 & 313 & 1.06 & 6.35 & 94.89 & 0.07 \\
 & 768$\times$7$\times$7 & 3.38 & 88889 & 302 & 0.68 & 4.93 & 150.89 & 0.02 \\ \cline{2-9} 
 &\multicolumn{2}{c|}{\cc Average} &\cc - & \cc{313} &\cc - & \cc{6.42} &\cc - & \cc{0.12} \\ \midrule 
\multirow{5}{*}{\begin{tabular}[c]{@{}c@{}}PConv 3$\times$3\\ (ours, with $r=\frac{1}{4}$)
\end{tabular}} & 96$\times$56$\times$56 & 162 & 9215 & 1493 & 5.46 & 29.67 & 85.30 & 1.90 \\
 & 192$\times$28$\times$28 & 162 & 14360 & 2326 & 3.09 & 52.43 & 66.46 & 2.44 \\
 & 384$\times$14$\times$14 & 162 & 24408 & 3954 & 3.58 & 45.25 & 49.98 & 3.24 \\
 & 768$\times$7$\times$7 & 162 & 32866 & 5324 & 5.02 & 32.27 & 48.30 & 3.35 \\ \cline{2-9} 
 &\multicolumn{2}{c|}{\cc Average} &\cc - &\cc{3274} &\cc - &\cc {39.91} &\cc - &\cc {2.73} \\ \bottomrule
\end{tabular}%
}
\vspace{-0.08in}
\caption{On-device FLOPS for different operations. PConv appears as an appealing choice for high FLOPS with reduced FLOPs.}
\label{FLOPS_compare}
\vspace{-0.15in}
\end{table*}

%% file: tables/pconv_pwconv.tex
\begin{table}
\vspace{0.05in}
\centering
\resizebox{.9\linewidth}{!}{%
\setlength{\tabcolsep}{3pt}
\begin{tabular}{cccc}
\toprule
Stage & \small{DWConv+PWConv} & \begin{tabular}[c]{@{}c@{}}\small{GConv+PWConv} \\ (16 groups)\end{tabular} & \begin{tabular}[c]{@{}c@{}}\cc \small{PConv+PWConv}\\ \cc $r=\frac{1}{4}$\end{tabular} \\ \midrule
1       & 0.0089 & 0.0065 & \cc 0.0069          \\
2       & 0.0158 & 0.0137          & \cc{0.0136} \\
3       & 0.0214 & 0.0202          & \cc{0.0172} \\
4       & 0.0130 & 0.0128          & \cc{0.0115} \\ \midrule
Average & 0.0148 & 0.0133          & \cc{0.0123} \\ \bottomrule
\end{tabular}%
}
\vspace{-0.05in}
\caption{A PConv followed by a PWConv well approximates the regular Conv $3 \times 3$ at different stages of a pre-trained ResNet50. PConv + PWConv together have the lowest test loss on average.}
\label{tab:PConvPWConv}
\vspace{-0.15in}
\end{table}

%% file: tables/imagenet.tex
\begin{table}
\vspace{-0.05in}
\centering
\resizebox{1\linewidth}{!}{%
\setlength{\tabcolsep}{0.01pt}
\begin{tabular}{@{\hskip -0.01in}l@{\hskip -0.08in}ccccccc}
\toprule
    \multicolumn{1}{c}{{Network}} &
  { \begin{tabular}[c]{c}Params\\(M)\end{tabular}} &
  { \begin{tabular}[c]{c}\small FLOPs\\(G)\end{tabular}} &
  { \begin{tabular}[c]{c}\small Throughput\\on GPU\\(fps) $\uparrow$\end{tabular}} &
  { \begin{tabular}[c]{c}\small Latency\\ on CPU\\ (ms) $\downarrow$\end{tabular}} &
  { \begin{tabular}[c]{c}\small Latency\\ on ARM \\ (ms) $\downarrow$\end{tabular}} &
  { \begin{tabular}[c]{c}Acc. \\ (\%)\end{tabular}} \\ \midrule
{ShuffleNetV2 $\times$1.5~\cite{ma2018shufflenet}} &
  {3.5} &
  {0.30} &
  {4878} &
  {12.1} &
  {266} &
  {72.6} \\
{MobileNetV2~\cite{sandler2018mobilenetv2}} &
  {3.5} &
  {0.31} &
  {4198} &
  {12.2} &
  {442} &
  {72.0} \\
{MobileViT-XXS~\cite{mehta2021mobilevit}} &
  {1.3} &
  {0.42} &
  {2393} &
  {30.8} &
  {348} &
  {69.0} \\
{EdgeNeXt-XXS~\cite{maaz2022edgenext}} &
  {1.3} &
  {0.26} &
  {2765} &
  {15.7} &
  {239} &
  {71.2} \\
{\cc FasterNet-T0} &
  {\cc 3.9} &
  {\cc 0.34} &
  {\cc 6807} &
  {\cc 9.2} &
  {\cc 143} &
  {\cc 71.9}
\\ \midrule
{GhostNet $\times$1.3~\cite{han2020ghostnet}} &
  {7.4} &
  {0.24} &
  {2988} &
  {17.9} &
  {481} &
  {75.7} \\
{ShuffleNetV2 $\times$2~\cite{ma2018shufflenet}} &
  {7.4} &
  {0.59} &
  {3339} &
  {17.8} &
  {403} &
  {74.9} \\
{MobileNetV2 $\times$1.4~\cite{sandler2018mobilenetv2}} &
  {6.1} &
  {0.60} &
  {2711} &
  {22.6} &
  {650} &
  {74.7} \\
{MobileViT-XS~\cite{mehta2021mobilevit}} &
  {2.3} &
  {1.05} &
  {1392} &
  {40.8} &
  {648} &
  {74.8} \\
{EdgeNeXt-XS~\cite{maaz2022edgenext}} &
  {2.3} &
  {0.54} &
  {1738} &
  {24.4} &
  {434} &
  {75.0} \\
{PVT-Tiny~\cite{wang2021pyramid}} &
  {13.2} &
  {1.94} &
  {1266} &
  {55.6} &
  {708} &
  {75.1} \\
{\cc FasterNet-T1} &
  {\cc 7.6} &
  {\cc 0.85} &
  {\cc 3782} &
  {\cc 17.7} &
  {\cc 285} &
  {\cc 76.2} \\ \midrule
{CycleMLP-B1~\cite{chen2021cyclemlp}} &
  {15.2} &
  {2.10} &
  {865} &
  {116.1} &
  {892} &
  {79.1} \\
{PoolFormer-S12~\cite{yu2022metaformer}} &
  {11.9} &
  {1.82} &
  {1439} &
  {49.0} &
  {665} &
  {77.2} \\
{MobileViT-S~\cite{mehta2021mobilevit}} &
  {5.6} &
  {2.03} &
  {1039} &
  {56.7} &
  {941} &
  {78.4} \\
{EdgeNeXt-S~\cite{maaz2022edgenext}} &
  {5.6} &
  {1.26} &
  {1128} &
  {39.2} &
  {743} &
  {79.4} \\
{ResNet50~\cite{he2016deep,liu2022convnet}} &
  {25.6} &
  {4.11} &
  {959} &
  {73.0} &
  {1131} &
  {78.8} \\
{\cc FasterNet-T2} &
  {\cc 15.0} &
  {\cc 1.91} &
  {\cc 1991} &
  {\cc 33.5} &
  {\cc 497} &
  {\cc 78.9} \\ \midrule
{CycleMLP-B2~\cite{chen2021cyclemlp}} &
  {26.8} &
  {3.90} &
  {528} &
  {186.3} &
  {1502} &
  {81.6} \\
{PoolFormer-S24~\cite{yu2022metaformer}} &
  {21.4} &
  {3.41} &
  {748} &
  {92.8} &
  {1261} &
  {80.3} \\
{PoolFormer-S36~\cite{yu2022metaformer}} &
  {30.9} &
  {5.00} &
  {507} &
  {138.0} &
  {1860} &
  {81.4} \\
{ConvNeXt-T~\cite{liu2022convnet}} &
  {28.6} &
  {4.47} &
  {657} &
  {86.3} &
  {1889} &
  {82.1} \\
{Swin-T~\cite{liu2021swin}} &
  {28.3} &
  {4.51} &
  {609} &
  {122.2} &
  {1424} &
  {81.3} \\
{PVT-Small~\cite{wang2021pyramid}} &
  {24.5} &
  {3.83} &
  {689} &
  {89.6} &
  {1345} &
  {79.8} \\
{PVT-Medium~\cite{wang2021pyramid}} &
  {44.2} &
  {6.69} &
  {438} &
  {143.6} &
  {2142} &
  {81.2} \\
{\cc FasterNet-S} &
  {\cc 31.1} &
  {\cc 4.56} &
  {\cc 1029} &
  {\cc 71.2} &
  {\cc 1103} &
  {\cc 81.3} \\ \midrule
{PoolFormer-M36~\cite{yu2022metaformer}} &
  {56.2} &
  {8.80} &
  {320} &
  {215.0} &
  {2979} &
  {82.1} \\
{ConvNeXt-S~\cite{liu2022convnet}} &
  {50.2} &
  {8.71} &
  {377} &
  {153.2} &
  {3484} &
  {83.1} \\
{Swin-S~\cite{liu2021swin}} &
  {49.6} &
  {8.77} &
  {348} &
  {224.2} &
  {2613} &
  {83.0} \\
{PVT-Large~\cite{wang2021pyramid}} &
  {61.4} &
  {9.85} &
  {306} &
  {203.4} &
  {3101} &
  {81.7} \\
{\cc FasterNet-M} &
  {\cc 53.5} &
  {\cc 8.74} &
  {\cc 500} &
  {\cc 129.5} &
  {\cc 2092} &
  {\cc 83.0} \\ \midrule
{PoolFormer-M48~\cite{yu2022metaformer}} &
  {73.5} &
  {11.59} &
  {242} &
  {281.8} &
  {OOM} &
  {82.5} \\
{ConvNeXt-B~\cite{liu2022convnet}} &
  {88.6} &
  {15.38} &
  {253} &
  {257.1} &
  {OOM} &
  {83.8} \\
{Swin-B~\cite{liu2021swin}} &
  {87.8} &
  {15.47} &
  {237} &
  {349.2} &
  {OOM} &
  {83.5} \\
{\cc FasterNet-L} &
  {\cc 93.5} &
  {\cc 15.52} &
  {\cc 323} &
  {\cc 219.5} &
  {\cc OOM} &
  {\cc 83.5} \\ \bottomrule
\end{tabular}%
}
\vspace{-0.1in}
\caption{Comparison on ImageNet-1k benchmark. Models with similar top-1 accuracy are grouped together. For each group, our FasterNet achieves the highest throughput on GPU and the lowest latency on CPU and ARM. 
All models are evaluated at $224\times224$ resolution except for the MobileViT and EdgeNeXt with $256\times256$. 
OOM is short for out of memory.}
\label{tab:imagenet}
\vspace{-0.1in}
\end{table}

%% file: tables/coco.tex
\begin{table*}
\centering
\resizebox{.9\linewidth}{!}{%
\setlength{\tabcolsep}{5pt}
\begin{tabular}{lccccccccc}
\toprule
Backbone &
\begin{tabular}[c]{c}Params \\ (M)\end{tabular} &
\begin{tabular}[c]{c}FLOPs \\ (G)\end{tabular} &
\begin{tabular}[c]{c}Latency on\\GPU (ms)\end{tabular} &
\hspace{0.1cm} $AP^b$ \hspace{0.1cm}&
\hspace{0.1cm} $AP^b_{50}$ \hspace{0.1cm}&
\hspace{0.1cm} $AP^b_{75}$ \hspace{0.1cm}&
\hspace{0.1cm} $AP^m$ \hspace{0.1cm}&
\hspace{0.1cm} $AP^m_{50}$ \hspace{0.1cm}&
\hspace{0.1cm} $AP^m_{75}$ \hspace{0.1cm} \\ \midrule
ResNet50~\cite{he2016deep}        & 44.2  & 253   & 54.9  & 38.0 & 58.6 & 41.4 & 34.4 & 55.1 & 36.7 \\
PoolFormer-S24~\cite{yu2022metaformer}    & 41.0  & 233   & 111.0 & 40.1 & 62.2 & 43.4 & 37.0 & 59.1 & 39.6 \\
PVT-Small~\cite{wang2021pyramid}        & 44.1  & 238   & 89.5  & 40.4 & 62.9 & 43.8 & 37.8 & 60.1 & 40.3 \\
\cc FasterNet-S      &\cc 49.0  &\cc 258   &\cc 54.3  &\cc 39.9 &\cc 61.2 &\cc 43.6 &\cc 36.9 &\cc 58.1 &\cc 39.7 \\ \midrule
ResNet101~\cite{he2016deep}       & 63.2  & 329   & 68.9  & 40.4 & 61.1 & 44.2 & 36.4 & 57.7 & 38.8 \\
ResNeXt101-32$\times$4d~\cite{xie2017aggregated} & 62.8  & 333   & 80.5  & 41.9 & 62.5 & 45.9 & 37.5 & 59.4 & 40.2 \\
PoolFormer-S36~\cite{yu2022metaformer}    & 50.5  & 266   & 146.9 & 41.0 & 63.1 & 44.8 & 37.7 & 60.1 & 40.0 \\
PVT-Medium~\cite{wang2021pyramid}       & 63.9  & 295   & 117.3 & 42.0 & 64.4 & 45.6 & 39.0 & 61.6 & 42.1 \\
\cc FasterNet-M      &\cc 71.2  &\cc 344   &\cc 71.4  &\cc 43.0 &\cc 64.4 &\cc 47.4 &\cc 39.1 &\cc 61.5 &\cc 42.3 \\ \midrule
ResNeXt101-64$\times$4d~\cite{xie2017aggregated} & 101.9 & 487   & 112.9 & 42.8 & 63.8 & 47.3 & 38.4 & 60.6 & 41.3 \\
PVT-Large~\cite{wang2021pyramid}        & 81.0  & 358   & 152.2 & 42.9 & 65.0 & 46.6 & 39.5 & 61.9 & 42.5 \\
\cc FasterNet-L      &\cc 110.9 &\cc 484 &\cc 93.8  &\cc 44.0 &\cc 65.6 &\cc 48.2 &\cc 39.9 &\cc 62.3 &\cc 43.0 \\ \bottomrule
\end{tabular}%
}
\vspace{-0.08in}
\caption{Results on COCO object detection and instance segmentation benchmarks. FLOPs are calculated with image size (1280, 800).}
\label{tab:coco}
\vspace{-0.1in}
\end{table*}

%% file: tables/ablation.tex
\begin{table}
\centering
\resizebox{1.\linewidth}{!}{%
\setlength{\tabcolsep}{1pt}
\begin{tabular}{llcccc}
\toprule
Ablation&
Variant &
\begin{tabular}[c]{@{}c@{}} {\small Throughput}  \\ on {\small GPU} \\ (fps)\end{tabular} &
\begin{tabular}[c]{@{}c@{}} {\small Latency} \\ on {\small CPU} \\ (ms)\end{tabular} &
\begin{tabular}[c]{@{}c@{}} {\small Latency} \\ on {\small ARM} \\ (ms)\end{tabular} &
\begin{tabular}[c]{@{}c@{}}Acc. \\ (\%)\end{tabular} \\ \midrule
\multirow{3}{*}{Partial ratio} & $\text{T0}^*$ w/  $r = 1/2$   &6626       & 9.6       & 145       & 71.7      \\
                               &\cc T0 w/ $r = 1/4$ &\cc 6807   &\cc 9.2    &\cc 143    &\cc 71.9   \\
                               & $\text{T0}^*$ w/ $r = 1/8$    &6204	     &8.9        &140        & 71.3    \\ \midrule
\multirow{2}{*}{Normalization} &\cc T0 w/ BN          &\cc 6807   &\cc 9.2    &\cc 143 &\cc 71.9                   \\
                               & T0 w/ LN          &5515	  &10.7 &159                         & 71.9                     \\ \midrule
\multirow{5}{*}{Activation}    & T0 w/ ReLU        &6929	  &8.2  &114                         & 71.3                     \\
                               &$\text{T0}^*$ w/ ReLU        & 5866                 & 9.3                  & 143 & 71.7 \\
                               &\cc T0 w/ GELU        &\cc 6807   &\cc 9.2    &\cc 143 &\cc 71.9                     \\
                               &\cc T2 w/ ReLU        &\cc 1991                 &\cc 33.5                 &\cc 497 &\cc 78.9                     \\
                               & T2 w/ GELU        &1985	  &35.4  &557                         & 78.7 \\ 
\bottomrule
\end{tabular}%
}
\vspace{-0.1in}
\caption{Ablation on the partial ratio, normalization, and activation of FasterNet. Rows highlighted in grey are the default settings. $\text{T0}^*$ denotes T0 variants with modified network width and depth.}
\label{tab:ablation}
\vspace{-0.1in}
\end{table}

%% file: 5_conclusion.tex
\section{Conclusion}
\label{sec:conclusion}
In this paper, we have investigated the common and unresolved issue that many established neural networks suffer from low floating-point operations per second (FLOPS). We have revisited a bottleneck operator, DWConv, and analyzed its main cause for a slowdown -- frequent memory access. To overcome the issue and achieve faster neural networks, we have proposed a simple yet fast and effective operator, PConv, that can be readily plugged into many existing networks. We have further introduced our general-purpose FasterNet, built upon our PConv, that achieves state-of-the-art speed and accuracy trade-off on various devices and vision tasks. We hope that our PConv and FasterNet would inspire more research on simple yet effective neural networks, going beyond academia to impact the industry and community directly.

%% file: 6_appendix.tex
\appendix
\addcontentsline{toc}{section}{Appendix}
\section*{\Large Appendix}
In this appendix, we provide further details on the experimental settings, full comparison plots, architectural configurations, PConv implementations, comparisons with related work, limitations, and future work.

\section{ImageNet-1k experimental settings}
\label{imagenet_settings}
We provide ImageNet-1k training and evaluation settings in~\cref{tab:imagenet_settings}. They can be used for reproducing our main results in~\cref{tab:imagenet} and \cref{fig:imageent}. Different FasterNet variants vary in the magnitude of regularization and augmentation techniques. The magnitude increases as the model becomes larger to alleviate overfitting and improve accuracy. Note that most of the compared works in~\cref{tab:imagenet} and \cref{fig:imageent}, \eg, MobileViT, EdgeNext, PVT, CycleMLP, ConvNeXt, Swin, \etc, also adopt such advanced training techniques (ADT). Some even heavily rely on the hyper-parameter search. For others w/o ADT, \ie, ShuffleNetV2, MobileNetV2, and GhostNet, though the comparison is not totally fair, we include them for reference.

\section{Downstream tasks experimental settings}
For object detection and instance segmentation on the COCO2017 dataset, we equip our FasterNet backbone with the popular Mask R-CNN detector. 
We use ImageNet-1k pre-trained weights to initialize the backbone and Xavier to initialize the add-on layers. Detailed settings are summarized in~\cref{tab:coco_settings}.

\section{Full comparison plots on ImageNet-1k}
\begin{figure*}
    \centering
    \includegraphics[width=1.\linewidth]{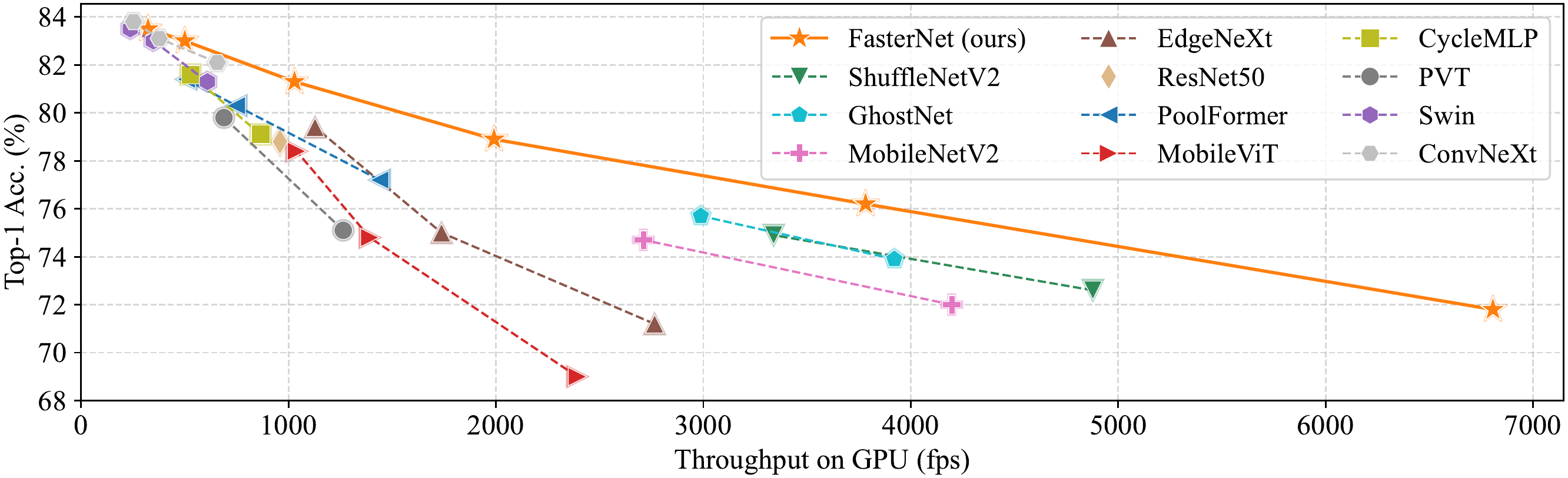}

    \includegraphics[width=1.\linewidth]{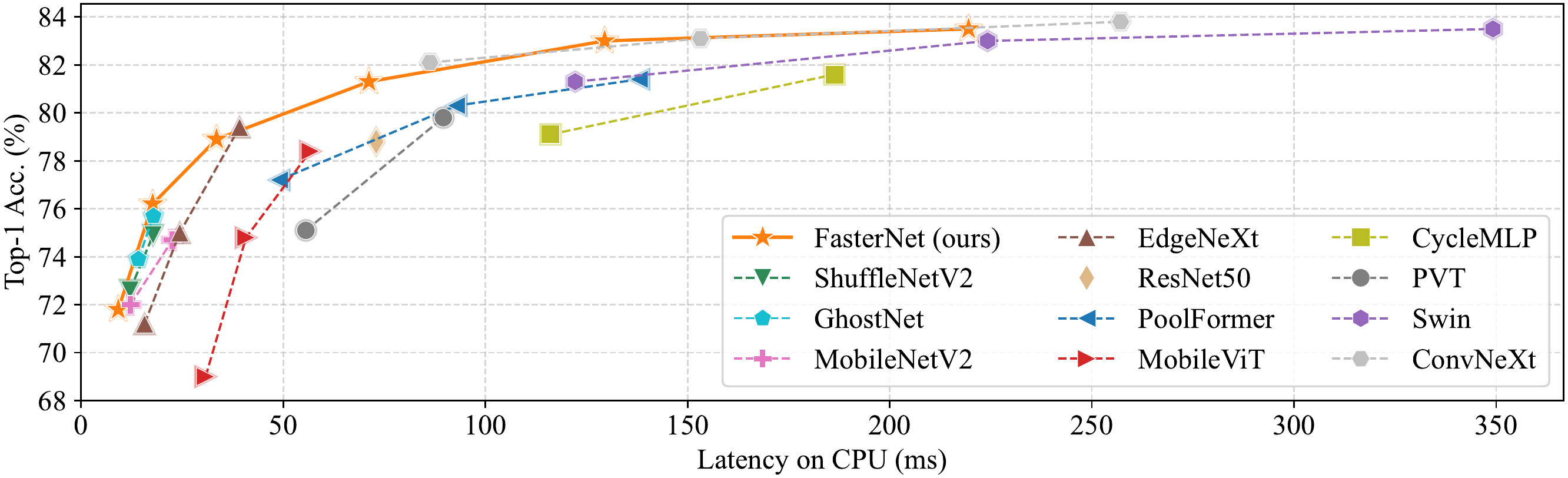}
    
    \includegraphics[width=1.\linewidth]{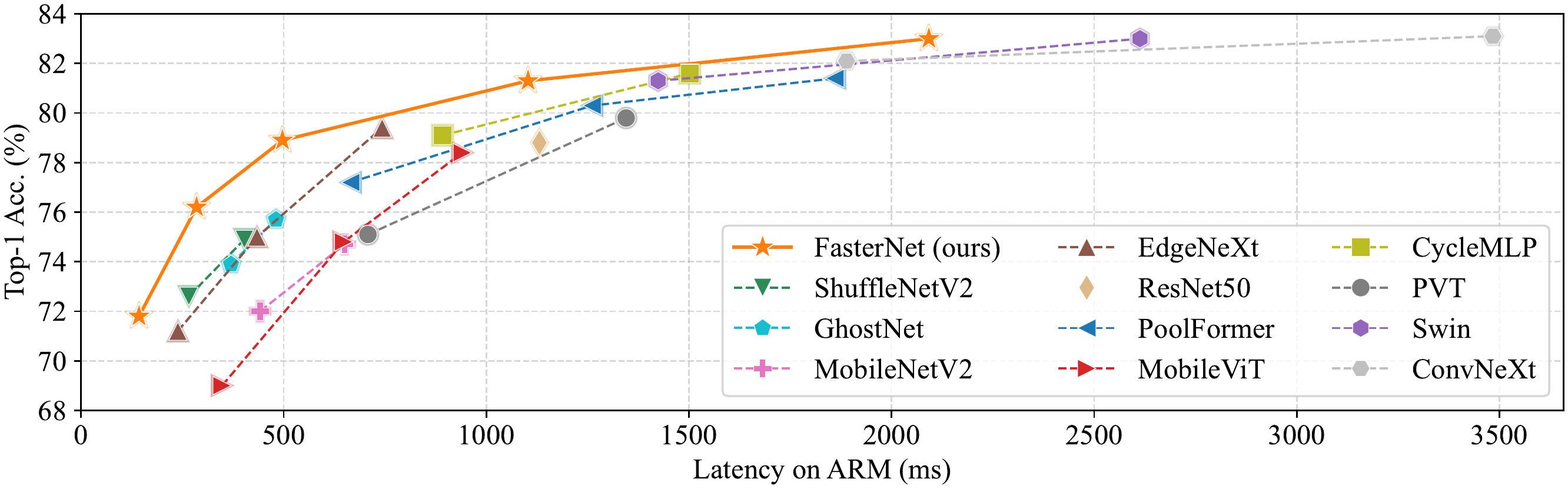}
    
    \vspace{-0.1in}
    \caption{Comparison of FasterNet with state-of-the-art networks. FasterNet consistently achieves better accuracy-throughput (the top plot) and accuracy-latency (the medium and bottom plots) trade-offs than others.}
    \label{fig:imagenet_full}
\end{figure*}

\cref{fig:imagenet_full} shows the full comparison plots on ImageNet-1k, which is the extension of~\cref{fig:imageent} in the main paper with a larger range of latency. \cref{fig:imagenet_full} shows consistent results that FasterNet strikes better trade-offs than others in balancing accuracy and latency/throughput on GPU, CPU, and ARM processors.

\section{Detailed architectural configurations}
We present the detailed architectural configurations in~\cref{tab:configuration}. While different FasterNet variants share a unified architecture, they vary in the network width (the number of channels) and network depth (the number of FasterNet blocks at each stage). The classifier at the end of the architecture is used for classification tasks but removed for other downstream tasks.

\input{tables/training_recipe}
\input{tables/coco_settings.tex}


\section{More comparisons with related work}
\medskip\noindent\textbf{Improving FLOPS.} \enspace 
There are a few other works~\cite{ding2022scaling_short,xia2022trt_short} also looking into the FLOPS issue and trying to improve it. They generally follow existing operators and try to find their proper configurations, \eg, RepLKNet~\cite{ding2022scaling_short} simply increases the kernel size while TRT-ViT~\cite{xia2022trt_short} reorders different blocks in the architecture. By contrast, this paper advances the field by proposing a novel and efficient PConv, opening up new directions and potentially larger room for FLOPS improvement.

\input{tables/configuration.tex}

\medskip\noindent\textbf{PConv vs. GConv.} \enspace 
PConv is schematically equivalent to a modified GConv~\cite{krizhevsky2012imagenet} that operates on a single group and leaves other groups untouched. Though simple, such a modification remains unexplored before. It's also significant in the sense that it prevents the operator from excessive memory access and is computationally more efficient. From the perspective of low-rank approximations, PConv improves GConv by further reducing the intra-filter redundancy beyond the inter-filter redundancy~\cite{haase2020rethinking_short}.

\medskip\noindent\textbf{FasterNet vs. ConvNeXt.} \enspace 
Our FasterNet appears similar to ConvNeXt~\cite{liu2022convnet} after substituting DWConv with our PConv. However, they are different in motivations. While ConvNeXt searches for a better structure by trial and error, we append PWConv after PConv to better aggregate information from all channels. Moreover, ConvNeXt follows ViT to use fewer activation functions, while we intentionally remove them from the middle of PConv and PWConv, to minimize their error in approximating a regular Conv. 

\medskip\noindent\textbf{Other paradigms for efficient inference.} \enspace 
Our work focuses on efficient network design, orthogonal to the other paradigms, \eg, neural architecture search (NAS)~\cite{elsken2019neural}, network pruning~\cite{molchanov2016pruning}, and knowledge distillation~\cite{hinton2015distilling}. They can be applied in this paper for better performance. However, we opt not to do so to keep our core idea centered and to make the performance gain clear and fair.

\medskip\noindent\textbf{Other partial/masked convolution works.} \enspace 
There are several works~\cite{liu2018image,gao2022convmae,liu2022partial} sharing similar names with our PConv. However, they differ a lot in objectives and methods. For example, they apply filters on partial pixels to exclude invalid patches~\cite{liu2018image}, enable self-supervised learning~\cite{gao2022convmae}, or synthesize novel images~\cite{liu2022partial}, while we target at the channel dimension for efficient inference. 

\section{Limitations and future work}
We have demonstrated that PConv and FasterNet are fast and effective, being competitive with existing operators and networks. Yet there are some minor technical limitations of this paper. For one thing, PConv is designed to apply a regular convolution on only a part of the input channels while leaving the remaining ones untouched. Thus, the stride of the partial convolution should always be 1, in order to align the spatial resolution of the convolutional output and that of the untouched channels. Note that it is still feasible to down-sample the spatial resolution as there can be additional downsampling layers in the architecture. 
And for another, our FasterNet is simply built upon convolutional operators with a possibly limited receptive field. Future efforts can be made to enlarge its receptive field and combine it with other operators to pursue higher accuracy.


%% file: tables/training_recipe.tex
\begin{table}
\vspace{0.1in}
\centering
\resizebox{1\linewidth}{!}{%
\begin{tabular}{@{}l|cccccc@{}}
\toprule
Variants           & T0                         & T1    & T2    & S    & M     & L            \\ \midrule
Train Res          & \multicolumn{6}{c}{192 for epoch 1$\sim$280, 224 for epoch 281$\sim$300}                                    \\
Test Res           & \multicolumn{6}{c}{224}                                    \\ \midrule
Epochs             & \multicolumn{6}{c}{300}                                    \\
\# of forward pass & \multicolumn{6}{c}{188k}                                   \\ \midrule
Batch size         & 4096   & 4096      & 4096      & 4096      & 2048  & 2048\\
Optimizer          & \multicolumn{6}{c}{AdamW}                                  \\
Momentum & \multicolumn{6}{c}{0.9/0.999}                              \\
LR                 & 0.004  & 0.004 & 0.004 & 0.004 & 0.002 & 0.002             \\
LR decay           & \multicolumn{6}{c}{cosine}                                 \\
Weight decay       & 0.005  & 0.01  & 0.02  & 0.03  & 0.05 & 0.05               \\
Warmup epochs      & \multicolumn{6}{c}{20}                                     \\
Warmup schedule    & \multicolumn{6}{c}{linear}                                 \\ \midrule
Label smoothing    & \multicolumn{6}{c}{0.1}                                    \\
Dropout            & \multicolumn{6}{c}{{\color[HTML]{9B9B9B} \ding{55}}}       \\
Stoch. Depth       & {\color[HTML]{9B9B9B} \ding{55}}   & 0.02  & 0.05  & 0.1  & 0.2  & 0.3   \\
Repeated Aug       & \multicolumn{6}{c}{{\color[HTML]{9B9B9B} \ding{55}}}       \\
Gradient Clip.     & {\color[HTML]{9B9B9B} \ding{55}}   & {\color[HTML]{9B9B9B} \ding{55}}  & {\color[HTML]{9B9B9B} \ding{55}}  & {\color[HTML]{9B9B9B} \ding{55}}  & 1  & 0.01       \\ \midrule
H. flip            & \multicolumn{6}{c}{\ding{51}}                              \\
RRC                & \multicolumn{6}{c}{\ding{51}}                              \\
Rand Augment       & {\color[HTML]{9B9B9B} \ding{55}}   & 3/0.5 & 5/0.5 & 7/0.5 & 7/0.5 & 7/0.5 \\
Auto Augment       & \multicolumn{6}{c}{{\color[HTML]{9B9B9B} \ding{55}}}       \\
Mixup alpha        & 0.05                        & 0.1   & 0.1   & 0.3   & 0.5   & 0.7           \\
Cutmix alpha       & \multicolumn{6}{c}{1.0}                                    \\
Erasing prob. & \multicolumn{6}{c}{{\color[HTML]{9B9B9B} \ding{55}}} \\
Color Jitter       & \multicolumn{6}{c}{{\color[HTML]{9B9B9B} \ding{55}}}       \\
PCA lighting       & \multicolumn{6}{c}{{\color[HTML]{9B9B9B} \ding{55}}}       \\ \midrule
SWA                & \multicolumn{6}{c}{{\color[HTML]{9B9B9B} \ding{55}}}       \\
EMA                & \multicolumn{6}{c}{{\color[HTML]{9B9B9B} \ding{55}}}       \\ \midrule
Layer scale    & \multicolumn{6}{c}{{\color[HTML]{9B9B9B} \ding{55}}}                          \\ \midrule
CE loss            & \multicolumn{6}{c}{\ding{51}}                              \\
BCE loss           & \multicolumn{6}{c}{{\color[HTML]{9B9B9B} \ding{55}}}       \\ \midrule
Mixed precision    & \multicolumn{6}{c}{\ding{51}}                              \\ \midrule
Test crop ratio        & \multicolumn{6}{c}{0.9}       \\ \midrule
Top-1 acc. (\%)         & 71.9   & 76.2 & 78.9  & 81.3  & 83.0  &   83.5    \\ \bottomrule
\end{tabular}%
}
\caption{ImageNet-1k training and evaluation settings for different FasterNet variants.}
\label{tab:imagenet_settings}
\end{table}

%% file: tables/coco_settings.tex
\begin{table}
\centering
\vspace{0.2in}
\resizebox{\linewidth}{!}{%
\setlength{\tabcolsep}{6pt}
\begin{tabular}{@{}l|ccc@{}}
\toprule
Variants & \qquad S \qquad \qquad           & \qquad M \qquad \qquad      & L\\ \midrule
Train and test Res & \multicolumn{3}{c}{shorter side $=$ 800, longer side $\leq$ 1333} \\
Batch size                    & \multicolumn{3}{c}{16 (2 on each GPU)} \\
Optimizer                     & \multicolumn{3}{c}{AdamW}              \\
Train schedule     & \multicolumn{3}{c}{1$\times$ schedule (12 epochs)}                      \\
Weight decay                  & \multicolumn{3}{c}{0.0001}             \\
Warmup schedule               & \multicolumn{3}{c}{linear}             \\
Warmup iterations             & \multicolumn{3}{c}{500}                \\
LR decay           & \multicolumn{3}{c}{StepLR at epoch 8 and 11 with decay rate 0.1}             \\
LR                            & 0.0002      & 0.0001      & 0.0001     \\
Stoch. Depth                  & 0.15        & 0.2         & 0.3        \\ \bottomrule
\end{tabular}%
}
\caption{Experimental settings of object detection and instance segmentation on the COCO2017 dataset.}
\vspace{-0.2in}
\label{tab:coco_settings}
\end{table}

%% file: tables/configuration.tex
\begin{table*}
\centering
\resizebox{.93\linewidth}{!}{%
\begin{tabular}{@{}c|c|c|c|c|c|c|c|c|c@{}}
\toprule
Name &
  Output size &
  \multicolumn{2}{c|}{Layer specification} &
  T0 &
  T1 &
  T2 &
  S &
  M &
  L \\ \midrule
Embedding &
  \large{$\frac{h}{4} \times \frac{w}{4}$}&
  \begin{tabular}[c]{@{}c@{}}Conv\_4\_$c$\_4,\\ BN\end{tabular} &
  \# Channels $c$ &
  40 &
  64 &
  96 &
  128 &
  144 &
  192 \\ \midrule
Stage 1 &
  \large{$\frac{h}{4} \times \frac{w}{4}$}&
  $\left[ \text{\begin{tabular}[c]{@{}c@{}}PConv\_3\_$c$\_1\_1/4,\\ Conv\_1\_$2c$\_1,\\ BN, Acti,\\ Conv\_1\_$c$\_1\end{tabular}}  \right] \times b_1 $  &
  \# Blocks $b_1$ &
  1 &
  1 &
  1 &
  1 &
  3 &
  3 \\ \midrule
Merging &
  \large{$\frac{h}{8} \times \frac{w}{8}$}&
  \begin{tabular}[c]{@{}c@{}}Conv\_2\_$2c$\_2,\\ BN\end{tabular} &
  \# Channels $2c$ &
  80 &
  128 &
  192 &
  256 &
  288 &
  384 \\ \midrule
Stage 2 &
  \large{$\frac{h}{8} \times \frac{w}{8}$}&
  $\left[ \text{\begin{tabular}[c]{@{}c@{}}PConv\_3\_$2c$\_1\_1/4,\\ Conv\_1\_$4c$\_1,\\ BN, Acti,\\ Conv\_1\_$2c$\_1\end{tabular}}  \right] \times b_2 $  &
  \# Blocks $b_2$ &
  2 &
  2 &
  2 &
  2 &
  4 &
  4 \\ \midrule
Merging &
  \large{$\frac{h}{16} \times \frac{w}{16}$}&
  \begin{tabular}[c]{@{}c@{}}Conv\_2\_$4c$\_2,\\ BN\end{tabular} &
  \# Channels $4c$ &
  160 &
  256 &
  384 &
  512 &
  576 &
  768 \\ \midrule
Stage 3 &
  \large{$\frac{h}{16} \times \frac{w}{16}$}&
  $\left[ \text{\begin{tabular}[c]{@{}c@{}}PConv\_3\_$4c$\_1\_1/4,\\ Conv\_1\_$8c$\_1,\\ BN, Acti,\\ Conv\_1\_$4c$\_1\end{tabular}}  \right] \times b_3 $  &
  \# Blocks $b_3$ &
  8 &
  8 &
  8 &
  13 &
  18 &
  18 \\ \midrule
Merging &
  \large{$\frac{h}{32} \times \frac{w}{32}$}&
  \begin{tabular}[c]{@{}c@{}}Conv\_2\_$8c$\_2,\\ BN\end{tabular} &
  \# Channels $8c$ &
  320 &
  512 &
  768 &
  1024 &
  1152 &
  1536 \\ \midrule
Stage 4 &
  \large{$\frac{h}{32} \times \frac{w}{32}$}&
  $\left[ \text{\begin{tabular}[c]{@{}c@{}}PConv\_3\_$8c$\_1\_1/4,\\ Conv\_1\_$16c$\_1,\\ BN, Acti,\\ Conv\_1\_$8c$\_1\end{tabular}}  \right] \times b_4 $  &
  \# Blocks $b_4$ &
  2 &
  2 &
  2 &
  2 &
  3 &
  3 \\ \midrule
Classifier &
  $1  \times 1$ &
  \begin{tabular}[c]{@{}c@{}}Global average pool,\\ Conv\_1\_1280\_1,\\ Acti,\\ FC\_1000\end{tabular} &
  Acti &
  GELU &
  GELU &
  ReLU &
  ReLU &
  ReLU &
  ReLU \\ \midrule
\multicolumn{4}{c|}{FLOPs (G)} &
  0.34 &
  0.85 &
  1.90 &
  4.55 &
  8.72 &
  15.49 \\ \midrule
\multicolumn{4}{c|}{Params (M)} &
  3.9 &
  7.6 &
  15.0 &
  31.1 &
  53.5 &
  93.4 \\ \bottomrule
\end{tabular}%
}
\caption{Configurations of different FasterNet variants. ``Conv\_$k$\_$c$\_$s$'' means a convolutional layer with the kernel size of $k$, the output channels of $c$, and the stride of $s$. ``PConv$\_k\_c\_s\_r$'' means a partial convolution with an extra parameter, the partial ratio of $r$. ``FC\_1000'' means a fully connected layer with 1000 output channels. $h \times w$ is the input size while $b_i$ is the number of FasterNet blocks at stage $i$. The FLOPs are calculated given the input size of $224 \times 224$.}
\label{tab:configuration}
\end{table*}